# Using Phonemes in cascaded S2S translation pipeline

**Rene Pilz**  and  **Johannes Schneider**

Department of Computer Science & Information Systems
University of Liechtenstein, Liechtenstein
rene.pilz@uni.li, johannes.schneider@uni.li

## Abstract

This paper explores the idea of using phonemes as a textual representation within a conventional multilingual simultaneous speech-to-speech translation pipeline, as opposed to the traditional reliance on text-based language representations. To investigate this, we trained an open-source sequence-to-sequence model on the WMT17 dataset in two formats: one using standard textual representation and the other employing phonemic representation. The performance of both approaches was assessed using the BLEU metric. Our findings shows that the phonemic approach provides comparable quality but offers several advantages, including lower resource requirements or better suitability for low-resource languages.

## 1 Introduction

According to Wang et al. (2022) simultaneous speech-to-speech (S2S) translation systems play a crucial role in enabling real-time multilingual communication. Conventionally, these systems employ a pipeline consisting of Automatic Speech Recognition (ASR), text-to-text translation, and Text-to-Speech (TTS) synthesis, each step utilizing standardized textual representations of language. However, this traditional methodology inherently depends on the existence of official written language forms and requires extensive speech data for training ASR systems, posing significant challenges for under-resourced or endangered languages. Jiang, Ahmed, Carson-Berndsen, Cahill, and Way (2011) and Do, Coler, Dijkstra, and Klabbers (2022) successfully utilized phonetic representations for under-resourced source languages. In particular, spoken dialects prevalent in many countries often lack an official written form.

Furthermore, recent progress in Text-to-Speech (TTS) models has shown a preference for phoneme-based embeddings, attributed to their improved performance. This typically involves an extra computational process where the text input is first converted to phonemes before being transformed into audio output.

To address these gaps, our study investigates whether adopting phonemic representations throughout the entire multilingual S2S pipeline provides distinct advantages. We trained an open-source sequence-to-sequence (seq2seq) model using the WMT17 dataset, comparing translation quality for English-to-German translations conducted at both textual and phonemic levels. Figure 2 shows the processes used to train and validate both models. By evaluating model outputs using standard BLEU scores we demonstrate that phoneme-based representations can offer comparable translation quality. Consequently, the phonemic approach might emerge as an advantageous alternative, particularly beneficial for under-resourced languages, while simultaneously streamlining the translation process by reducing computational complexity within the TTS step.

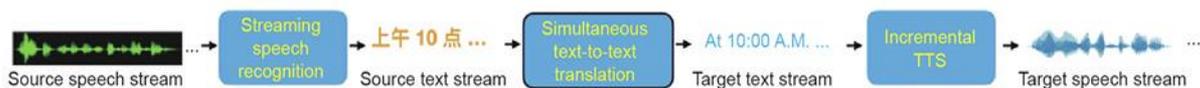

Figure 1: Pipeline of the cascaded S2S system (Wang, Wu, He, Huang, & Church, 2022)

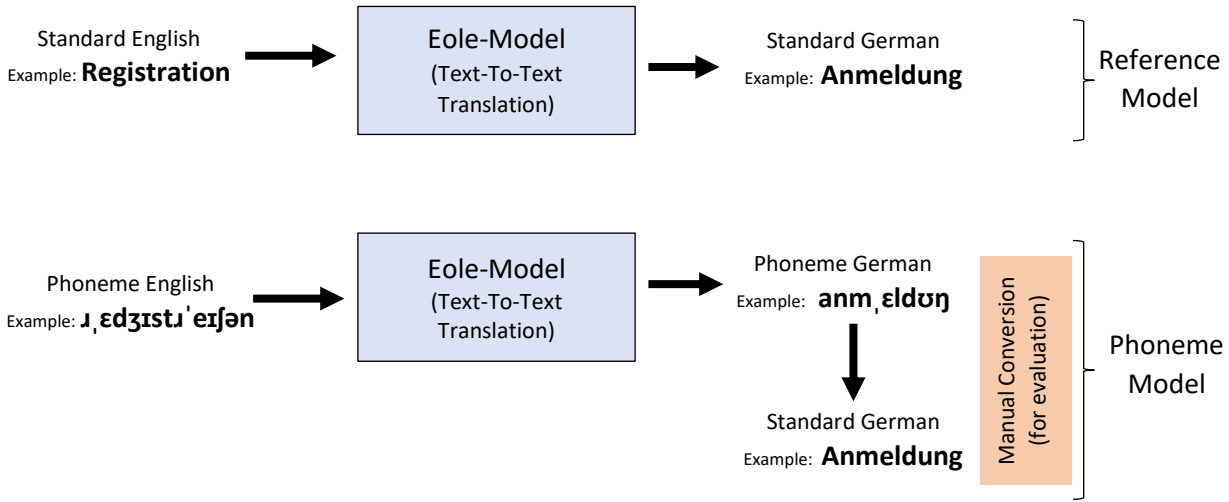

Figure 2: Reference and Phoneme Model

**Contribution** This study demonstrates that a sequence-to-sequence (seq2seq) model is capable of translating from English to German operating at the phonemic level with comparable quality measured using the BLEU score. Thus, phoneme-based translation might be preferable to conventional translation in particular in case of data scarcity and limited computational resources.

## 2 Related work

Wang et al. (2022) describe a prevalent methodology in AI-driven multilingual translation, which employs a simultaneous speech-to-speech (S2S) translation pipeline as shown in Figure 1. The process involves the use of standardized text representations of languages, adhering to the common strategy of decomposing complex problems into manageable sub-tasks:
1. Transcription of spoken language into written text.
2. Translation of text across languages.
3. Generation of spoken language from written text.

This segmentation facilitates individual analysis of each step, contingent upon the availability of standardized language text but also has limitations, i.e., the need for written language forms and extensive speech data.

The literature explored the substitution of the source language with phonetic representations but no approach replaced source and destination language by their phonetic representations.

Investigations into the latest TTS models reveal a predominant use of phoneme embedding, particularly in models based on advancements such as FastSpeech 2 (Ren et al., 2020), models based on StyleTTS 2 (Y. A. Li, Han, Raghavan, Mischler, & Mesgarani, 2023), models based on Transformer-TTS (N. Li et al., 2018), and models based on VITS (Kim, Kong, & Son, 2021). While models like Tacotron 2 (Shen et al., 2017) can operate on standard text representations, they generally exhibit improved performance with phoneme representation.

In essence, employing a TTS model reliant on phonemes necessitates the conversion of standard text into phoneme representation—an additional computational step that incurs extra processing time.

Gupta and Kumar (2021) demonstrated that seq2seq models are capable of translating between languages in text format, showcasing the versatility of these models in handling text-based translations. Therefore, this study also utilizes a seq2seq model for the translation process.

## 3 Materials and Methods

### 3.1 Model and Tools

The Eole model[1], a derivative of the OpenNMT Toolkit as developed by Klein, Kim, Deng, Senellart, and Rush (2017), was employed as the sequence-to-sequence (Seq2Seq) framework for translating English text or phonemes into German. To convert language text in German and English to phenom representation the espeak-ng[2] Framework was used. This framework is also used by the previously stated TTS models Tacotron 2 and FastSpeech 2.

---
[1] https://github.com/eole-nlp/eole
[2] https://github.com/espeak-ng/espeak-ng

| Model | Source sentence | Target sentence |
|---|---|---|
| **Reference** | Registration for the event can be submitted. | Die Anmeldung zur Veranstaltung kann vorgenommen werden. |
| **Phoneme** | ɹˌɛdʒɪstɹˈeɪʃən fəðɪ ɪvˈɛnt kan biː səbmˈɪtɪd | diː ˈanmˌɛldʊŋ tsuːɾ fɛrˈanʃtˌaltʊŋ kˌan fˈoːrɡənˌomən vˌɛrdən |

Table 1: Example (source, target) pair for training of the reference and phoneme model

### 3.2 Dataset

We performed a German-English news translation task, using the WMT17 dataset from the official website. The validation phase utilized the newstest2016 files for assessing performance[3].

For the development of our Phoneme Model, both the English source and the German target components of the WMT17 dataset were transformed into their phonetic equivalents (see examples in Table 1).

### 3.3 Measures

The BLEU score, introduced by Papineni, Roukos, Ward, and Zhu (2001), serves as a conventional tool for evaluating the quality of language translations. But this method is not suitable for comparing sentences written in phoneme format as words may exhibit acoustic similarity and multiple similar phonemic representations, as illustrated in Table 2. It is non-trivial and beyond the scope of this research to automatically create a standardized phonemic representation or match reliably similar phonemic representations. We decided to manually convert the output of the Phoneme Model to Standard German for evaluation (as shown in Figure 2). That is, we converted 100 sentences with a total amount of 2155 words from phonemic representation back to standard German, which are publicly available for reproducibility at https://github.com/fungus75/Phonemes_S2S_Pipeline . Due to the absence of phonemic representations for punctuation marks such as commas and full stops, these characters were eliminated prior to validation.

## 4 Experimental Setup

In our study, we juxtapose our phoneme-based methodology with a reference translation that adheres to the unaltered WMT17 recipe [4]. The configuration of this comparison is depicted in Figure 2. Initially, we trained the reference model and computed its BLEU score. Subsequently, we employed the identical recipe to train the Phoneme Model, albeit exclusively utilizing the phoneme dataset. Both models were trained from scratch on the same hardware (a Nvidia 3090 GPU) in about 24 hours. No modifications nor training parameter adjustments to the default WMT17 recipe of the Eole model were made.

## 5 Results

Table 3 presents a comparison between the two models. Both models are of comparable quality as measured by the BLEU score. A higher score correlates with a better translation outcome. Lavie (2011) posits that a BLEU score of 30 and above signifies that a translation is "understandable".

| Model | BLEU score |
|---|---|
| Reference Model | 39 |
| Phoneme Model | 38 |

Table 3: Benchmarking both models

|  | **Word 1** | **Word 2** | **Word 3** |
|---|---|---|---|
| Reference | ʊmzɛtsʊŋ | nɪçt | yːbɜ |
| Phoneme Model Output | ˈʊmzˌɛtsʊŋɡ | nˈɪçt   or   nˈiçt | ˌyːbɜ  or  ˌyːba |
| Standard German | Umsetzung | nicht | über |

Table 2: Different phonemic variants.

---

[3] https://www.statmt.org/wmt17/translation-task.html

[4] https://github.com/eole-nlp/eole/tree/main/recipes/wmt17

We observed that both models tend to exhibit enhanced performance on shorter sentences, as evidenced by higher scores. For instance, the sentence "She and her mother were absolutely best friends" was translated by both models perfectly as "Sie und ihre Mutter waren absolut beste Freunde". Shorter sentences might be more common and simpler in their structure facilitating translation.

Both models generally demonstrate a high proficiency in identifying personal names (e.g., Obama, Amy, Lynn Buford…); however, the models tend to lack semantic understanding of text. For instance, (English) names like "Miller" and "Tailor" are typically not translated to "Müller" and "Schneider". But we observed that "Professor Lamb" was frequently mistranslated as "Professor Lamm" (the German equivalent of "lamb"). Another example demonstrating that models lack semantic understanding is the following: Models exhibit difficulties in handling sentences with verbs that possess multiple meanings, particularly in brief sentences where context is limited. A notable example is the translation of "Obama receives Netanyahu", which was inaccurately rendered as "Obama erhält Netanjahu" interpreting "receives" as "get" rather than the intended "meets with". This tendency to default to the most common meaning of ambiguous verbs highlights a limitation in the models' contextual understanding. But because this issue was shown on both models – Reference and Phoneme – the limitation must be caused by the used Eole model rather than in the phoneme approach.

Comparative analysis of the outputs from the Reference Model and the Phoneme Model reveals distinct phrasing tendencies. For instance, the Reference Model output "Studenten sagten, sie würden sich auf seine Klasse freuen" contrasts with the Phoneme Model's prediction of "Studenten sagten, sie freuen sich auf seine Klasse." This observation suggests that the Phoneme Model may have a propensity to generate sentences in the present tense rather than the conditional mood.

## 6 Conclusion

This research illustrates that the development of the Eole sequence-to-sequence (seq2seq) model from scratch, utilizing a phonemic dataset, produces results that are on par with those achieved through the use of a traditional text dataset. Nonetheless, the adoption of a phonemic methodology provides several unique benefits:

1. Generating a phonemic representation of spoken language is simpler than producing standardized, official text, making phonemes particularly advantageous for under-resourced languages that may lack such standardized forms.
2. Specifically for Text-to-Speech (TTS) tasks, working directly with phonemes eliminates the necessity of translating standard text into phonemic representations. This benefit is not only crucial for under-resourced languages, which may lack established rules for such conversion, but it also streamlines any pipeline that transforms text into speech. Minimizing processing steps leads to reduced computation time, quicker outcomes, and diminished latency.

In conclusion, the integration of phonemic representations across the entire translation pipeline, not solely at the input stage, has the potential to yield outcomes of comparable quality in terms of BLEU score to those obtained through conventional methodologies. In particular, the present study concentrates on the exploration of the Eole seq2seq model, acknowledging its inherent constraints such as limited semantic understanding impacting both the reference and phoneme-based translation. Thus, our work can be seen as providing first evidence towards the benefits of phoneme-based translation. To establish the applicability of the proposed approach more broadly, subsequent research involving alternative models and more datasets is warranted. Furthermore, the application of XAI techniques might be beneficial to understand model behavior and possible strategies for improvements in more depth (Schneider, 2024). Leveraging strategies common in deep learning and LLMs like reflection or generation and refinement might also further improve model outcomes (Schneider, 2025; Schneider & Vlachos, 2024). Nonetheless, the findings of this investigation provide encouraging indications of the efficacy of the approach. On a larger scale, our work can contribute towards sustainable AI by reducing data needs (Schneider, Seidel, Basalla, & vom Brocke, 2023).

## Ethical Considerations and Limitations

To our knowledge, this research does not encompass any ethical concerns. We used the well-established WMT17 dataset and trained the Eole model from scratch, thus avoiding any privacy concerns.

A primary limitation identified during this study is the requirement for sequence-to-sequence (seq2seq) models to accommodate Unicode characters, a necessity stemming from the extensive employment of these characters in the representation of phonemes.